\documentclass[letterpaper]{article}
\usepackage[preprint]{aaai2027}
\usepackage[hyphens]{url}
\usepackage{graphicx}
\usepackage{natbib}
\usepackage{caption}
\usepackage{booktabs}
\usepackage{amsmath}
\usepackage{amssymb}
\usepackage{xcolor}
\usepackage{array}
\usepackage{tikz}
\usepackage{listings}

\definecolor{deltaBlueBackground}{RGB}{198,226,250}

\lstdefinestyle{rulebankentry}{
  basicstyle=\ttfamily\scriptsize\color{black},
  columns=fullflexible,
  keepspaces=true,
  showstringspaces=false,
  breaklines=true,
  breakatwhitespace=true,
  frame=single,
  framerule=0.3pt,
  rulecolor=\color{black!35},
  backgroundcolor=\color{black!2},
  framesep=4pt,
  xleftmargin=5pt,
  xrightmargin=5pt,
  aboveskip=4pt,
  belowskip=6pt,
  literate={≤}{{$\leq$}}1 {→}{{$\rightarrow$}}1
}

\newcommand{\markeddelta}[2]{\tikz[baseline=(delta.base)]{\node[fill=#1, text=black, rounded corners=1pt, inner xsep=0.9pt, inner ysep=0.5pt, outer sep=0pt] (delta) {#2};}}

\title{Learning a Rule Bank to Revise 3D CT Reports with Independent Evidence}
\author{
Renjie Liang, Mengxian Lyu, Zijian Xu, \\
Boxin Zhao, Yang Yang, Jinqian Pan, \\
Zhengkang Fan, Mei Liu, Yonghui Wu, Jie Xu
}
\affiliations{
University of Florida, Gainesville, FL, USA
}

\begin{document}
\maketitle

\begin{abstract}

Vision--language models can generate fluent radiology reports from 3D chest CT scans, yet they may still omit abnormalities or misstate findings. Independent image analysis using models separate from the report model provides additional observations, but acting on uncertain evidence may introduce new errors. We propose an agent system that learns a revision policy represented as an executable rule bank for each report model. Inspired by experience banks, the agent proposes and refines rules using revision outcomes across development cases. During inference, the retained rules guide local edits using independent classifier predictions and quantitative image measurements without retraining the report model. Unedited sentences are preserved, and each change is linked to the rule that triggered it. On CT-RATE, revision increases CE-F1 by 11.5 points for SelfCue-KD and 0.9 points for NV-Reason-CT, with gains across five draft-sampling seeds. SelfCue-KD's CT-FineBench score also increases by 15.2 points. Draft-only rule learning yields limited gains, while the two image-evidence sources provide complementary benefits for abnormality detection. Measurements also supply numerical details missing from the drafts. These results support learning model-specific rules to control how independent image evidence informs report revision. Code is available at \url{https://github.com/renjie-liang/CT-RuleBank}.

\end{abstract}

\begin{figure*}[t]
\centering
\includegraphics[width=\textwidth]{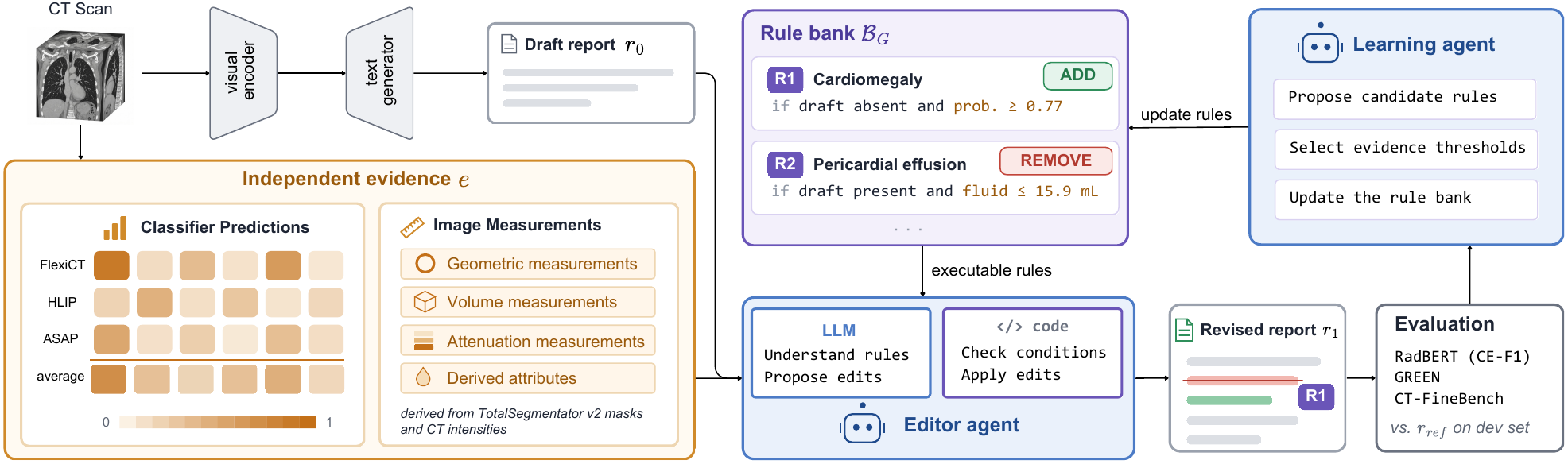}
\caption{Overview of rule bank learning and report revision. Independent evidence is obtained using image analysis models separate from the report model. The learning agent proposes rules and selects evidence thresholds using CE-F1 and GREEN feedback on development cases. At inference, the editor agent uses the fixed bank and independent evidence to make local edits to the draft.}
\label{fig:framework}
\end{figure*}

\section{Introduction}
\label{sec:intro}

Vision--language models can generate fluent radiology reports from 3D chest CT scans \citep{hamamci2026generalist,myronenko2026nv,wu2025towards}. Yet they can still miss important abnormalities or misstate findings that depend on precise image measurements \citep{liang2026selfcue,heiman2025factchexcker,hong2025diagnostic}. These errors motivate targeted revision: correcting erroneous statements and recovering omitted findings while preserving the correct content of the draft. In an agent-based revision workflow, a report model generates an initial draft from the CT scan, and an editor agent reviews and revises that draft.

An editor agent may rely on the draft and its own knowledge to identify errors and propose corrections. However, this approach provides no new observation of the scan, and intrinsic self-correction can be unreliable without external feedback \citep{huang2024large}. External feedback provides additional information to support more reliable revision \citep{gou2024critic,mahmood2025automatic}. Independent image analysis offers one source of such feedback, including predictions from a separate image classification model and quantitative measurements derived from image segmentation. A draft may describe a finding while omitting its numerical measurements. Independent image measurements provide values that can be added when the draft and evidence support the revision.

However, independent evidence can also be wrong. Image classifiers may misidentify abnormalities, and segmentation errors may affect quantitative measurements. Even an accurate measurement may have different diagnostic implications across patients. When the evidence disagrees with the draft, the editor agent therefore needs to decide whether the evidence justifies a change. Some report-generation methods incorporate diagnostic predictions into the generator \citep{jin2024promptmrg,di2025ct}. MedScribe uses tool evidence to refine reports, but leaves conflicts between the draft and the evidence to the final LLM's implicit judgment \citep{orlando2026medscribe}. This gap motivates learning an explicit policy for deciding when and how evidence should trigger a revision. This policy should also account for differences between report models: a change that corrects a common error in one model may be unnecessary or harmful for another.

Experience banks provide a way to learn such a policy from past revision outcomes \citep{zhao2024expel,ouyang2026reasoningbank,shen2026evo}. Methods such as ExpeL and ReasoningBank store textual lessons in the agent's prompt and rely on the LLM to apply them to new cases \citep{zhao2024expel,ouyang2026reasoningbank}. For report revision, these lessons need to describe the conditions under which a change is warranted. A correction that works for one scan may introduce an error in another. By evaluating candidate revisions against reference reports across development cases, the agent can learn when a change helps and when it causes harm. We represent this experience as executable rules that specify the conditions for a local edit, allowing their applicability to be checked explicitly on new cases.

We propose an agent system that learns a model-specific \textbf{rule bank} for revising 3D CT reports with independent image evidence without retraining the report model. During learning, the agent proposes candidate rules and refines them using feedback on both beneficial and harmful revisions across development cases. Each retained rule specifies which drafts are eligible, an optional condition on the image evidence, and a local edit. A separate bank is learned for each report model to account for its error patterns. During inference, the system checks the rules against the draft and the available evidence, then uses the applicable rules to guide local revisions. Content outside the selected edits is preserved, and each change is recorded together with its triggering rule and any evidence values used.

We evaluate the system on CT-RATE \citep{hamamci2026generalist} using SelfCue-KD and NV-Reason-CT \citep{liang2026selfcue,myronenko2026nv}, assessing both abnormality detection and report quality. Comparisons with draft-only rule learning examine the contribution of independent image evidence, while comparisons with direct LLM revision and fixed rules assess the value of learning a revision policy. We also examine how the minimum precision required for rule admission affects revision coverage, detection gains, and report quality. Further analyses assess sensitivity to the learning agent and input drafts. Reference-based edit assessments and illustrative cases examine revision quality and the evidence behind individual changes. Our contributions are as follows:
\begin{itemize}
  \item We propose an agent system that learns executable revision rules from outcomes across development cases. The rules are tailored to each report model and guide when and how independent image evidence is used to revise its drafts.
  \item We show that rule learning with independent image evidence yields larger abnormality-detection gains than learning from draft-derived signals alone, with complementary benefits from classifier predictions and quantitative measurements.
  \item We evaluate revision across report models and draft samples. We analyze the effects of rule-admission precision and assess edit quality through reference comparisons and case studies.
\end{itemize}

\section{Related work}
\label{sec:related}

\subsection{Medical Report Generation}
\label{sec:related-report-generation}
Vision--language models have been developed to generate radiology reports from chest radiographs and 3D CT scans. For 3D CT, CT-RATE provides paired scans and reports for training and evaluating report generation models \citep{hamamci2026generalist}. NV-Reason-CT combines native 3D visual encoding with radiologist-guided reasoning \citep{myronenko2026nv}, while SelfCue uses contrastive decoding to surface findings supported by the model's hidden states \citep{liang2026selfcue}. Some approaches use diagnostic predictions to guide report generation \citep{jin2024promptmrg,chen2025dia,di2025ct}. In contrast, our work focuses on revising an existing draft using independent evidence. Agent-based approaches such as MedScribe, RadAgent, and Radiologist Copilot use specialized tools to gather evidence and guide report generation \citep{orlando2026medscribe,roschewitz2026radagent,yu2025radiologist}. MARCH combines retrieval-assisted report revision with multi-agent discussion to resolve disagreements among proposed findings \citep{lin2026march}. MedErr-CT and RADAR evaluate whether systems can find, judge, and correct errors in CT reports \citep{kyung2025mederr,sun2026radar}. MedScribe \citep{orlando2026medscribe} acknowledges noisy tool evidence but does not explicitly handle conflicts with the draft, leaving them to the final LLM's implicit judgment. However, we use learned rules to explicitly control report revision.

\subsection{Experience Banks in Agent Systems}
\label{sec:related-experience-banks}
Experience banks store and organize lessons from past tasks to guide future agent decisions. Reflexion and ExpeL extract reusable lessons from task feedback and experience \citep{shinn2023reflexion,zhao2024expel}. ReasoningBank, Dynamic Cheatsheet, and ACE continually update memories and strategies as new tasks are completed \citep{ouyang2026reasoningbank,suzgun2026dynamic,zhang2026agentic}. Building on this idea, our rule bank represents experience as rules that explicitly control how independent evidence is used to revise a report. Voyager builds a library of executable skills through interaction with its environment \citep{wang2023voyager}, while AutoHarness uses environment feedback to synthesize and refine code that constrains agent actions \citep{lou2026autoharness}. GEPA reflects on execution traces to propose and test prompt updates \citep{agrawal2026gepa}, while Self-Harness identifies model-specific failures and validates proposed harness changes through regression testing \citep{zhang2026self}. In medical applications, Evo-MedAgent combines past clinical experiences with diagnostic rules and tool reliability estimates to guide diagnosis \citep{shen2026evo}. AgentScore uses LLM proposals and deterministic validation to construct clinical scoring rules \citep{estevez2026automatic}. In contrast, we learn revision rules for each report model from outcomes across cases and evaluate their effects on abnormality detection and report quality.

\section{Method}
\label{sec:method}

\begin{table*}[t]
\caption{Report generation and RuleBank revision on CT-RATE. RuleBank uses independent image evidence to revise drafts from SelfCue-KD and NV-Reason-CT. $\dagger$ denotes baselines evaluated under our protocol.}
\label{tab:main-performance}
\centering\footnotesize
\setlength{\tabcolsep}{2pt}
\renewcommand{\arraystretch}{1.12}
\newcommand{\mainmetric}[2]{#1\makebox[2.8em][l]{#2}}
\begin{tabular}{@{}p{0.25\textwidth}rrrrrrrr@{}}
\toprule
Model & \multicolumn{4}{c}{Clinical metrics} & \multicolumn{4}{c}{NLP metrics} \\
\cmidrule(lr){2-5}\cmidrule(lr){6-9}
 & \multicolumn{1}{c}{CE-F1} & \multicolumn{1}{c}{CRG} & \multicolumn{1}{c}{GREEN} & \multicolumn{1}{c}{CT-FineBench} & BLEU-1 & BLEU-4 & ROUGE-L & METEOR \\
\midrule
CT-CHAT \citep{hamamci2026generalist} & \mainmetric{18.4}{} & \mainmetric{36.8}{} & \mainmetric{35.8}{} & \mainmetric{15.8}{} & 38.1 & -- & 33.4 & 21.7 \\
BTB3D-16 \citep{hamamci2026better} & \mainmetric{25.8}{} & \mainmetric{37.0}{} & \mainmetric{--}{} & \mainmetric{--}{} & 43.9 & 21.3 & -- & 22.3 \\
OmniCT \citep{lin2026omnict} & \mainmetric{36.3}{} & \mainmetric{--}{} & \mainmetric{--}{} & \mainmetric{--}{} & -- & -- & -- & -- \\
U-VLM \citep{shi2026u} & \mainmetric{41.4}{} & \mainmetric{--}{} & \mainmetric{--}{} & \mainmetric{--}{} & 47.4 & 25.6 & -- & -- \\
CLarGen \citep{maye2026generating} & \mainmetric{48.6}{} & \mainmetric{47.2}{} & \mainmetric{--}{} & \mainmetric{--}{} & -- & 20.8 & 29.8 & 33.5 \\
$\mu^2$Tokenizer \citep{li2025mu} & \mainmetric{--}{} & \mainmetric{--}{} & \mainmetric{42.9}{} & \mainmetric{--}{} & -- & -- & -- & 35.9 \\
SliceWorld \citep{tian2026sliceworld} & \mainmetric{--}{} & \mainmetric{--}{} & \mainmetric{50.7}{} & \mainmetric{--}{} & 39.4 & 22.6 & 38.2 & 23.0 \\
SelfCue \citep{liang2026selfcue} & \mainmetric{48.1}{} & \mainmetric{--}{} & \mainmetric{51.0}{} & \mainmetric{--}{} & 42.4 & 22.1 & 35.6 & 21.0 \\
RadAgent \citep{roschewitz2026radagent} & \mainmetric{$\approx22$}{} & \mainmetric{--}{} & \mainmetric{--}{} & \mainmetric{--}{} & -- & -- & -- & -- \\
MedScribe-Template (8B) \citep{orlando2026medscribe} & \mainmetric{32.3}{} & \mainmetric{--}{} & \mainmetric{--}{} & \mainmetric{--}{} & 34.1 & -- & 18.3 & 33.6 \\
MedScribe-Template (70B) \citep{orlando2026medscribe} & \mainmetric{35.3}{} & \mainmetric{--}{} & \mainmetric{--}{} & \mainmetric{--}{} & 34.2 & -- & 18.5 & 33.4 \\
MARCH \citep{lin2026march} & \mainmetric{--}{} & \mainmetric{--}{} & \mainmetric{--}{} & \mainmetric{--}{} & 48.2 & 25.7 & 38.3 & 45.6 \\
\midrule
SelfCue-KD$^{\dagger}$ & \mainmetric{46.2}{} & \mainmetric{43.9}{} & \mainmetric{50.5}{} & \mainmetric{26.2}{} & 43.6 & 21.7 & 34.9 & 21.2 \\
SelfCue-KD + RuleBank & \mainmetric{57.7}{\textsuperscript{\markeddelta{deltaBlueBackground}{$\uparrow 11.5$}}} & \mainmetric{51.9}{\textsuperscript{\markeddelta{deltaBlueBackground}{$\uparrow 8.0$}}} & \mainmetric{49.8}{\textsuperscript{\markeddelta{red!12}{$\downarrow 0.7$}}} & \mainmetric{41.4}{\textsuperscript{\markeddelta{deltaBlueBackground}{$\uparrow 15.2$}}} & 45.6 & 22.4 & 34.6 & 21.7 \\
NV-Reason-CT$^{\dagger}$ & \mainmetric{63.0}{} & \mainmetric{54.0}{} & \mainmetric{45.8}{} & \mainmetric{44.9}{} & 36.6 & 12.5 & 21.9 & 17.4 \\
NV-Reason-CT + RuleBank & \mainmetric{63.9}{\textsuperscript{\markeddelta{deltaBlueBackground}{$\uparrow \phantom{0}0.9$}}} & \mainmetric{54.9}{\textsuperscript{\markeddelta{deltaBlueBackground}{$\uparrow 1.0$}}} & \mainmetric{45.9}{\textsuperscript{\markeddelta{deltaBlueBackground}{$\uparrow 0.2$}}} & \mainmetric{44.7}{\textsuperscript{\markeddelta{red!12}{$\downarrow \phantom{0}0.2$}}} & 36.7 & 12.5 & 21.9 & 17.4 \\
\bottomrule
\end{tabular}
\end{table*}

\subsection{Task and Overview}
\label{sec:overview}

Given a 3D chest CT scan $x$, a report model $G$ generates a draft report $r_0=G(x)$. Additional image analysis provides independent evidence $e$ from the same scan. Our goal is to produce a revised report $r_1$ that corrects errors in $r_0$ while preserving the statements that do not warrant revision. The evidence may be noisy, so a disagreement between $e$ and $r_0$ does not by itself justify a change. Our system learns a rule bank $\mathcal{B}_G$ for each report model (Figure~\ref{fig:framework}). The learning agent proposes and refines rules using feedback on revision outcomes across development cases. The retained rules specify the conditions under which evidence warrants a local revision. During inference, the editor agent applies the learned rules according to
\begin{equation}
    r_1 = \operatorname{Revise}(r_0,e;\mathcal{B}_G).
    \label{eq:report-revision}
\end{equation}
The revision trace records the triggering rule and any evidence values it uses.

\subsection{Independent Evidence}
\label{sec:evidence}

Independent evidence is defined as information obtained from additional image analysis performed independently of the report model. In this work, we use two types of independent evidence: classifier predictions and quantitative image measurements.

\paragraph{Classifier predictions.} Both report models evaluated in this work use COLIPRI-based visual encoders \citep{wald2026comprehensive,liang2026selfcue,myronenko2026nv}. We therefore obtain additional classifier evidence from separately pretrained encoders. Classification models are trained on FlexiCT, HLIP, and ASAP features \citep{li2026universal,zhao2025towards,wang2026asap}. These encoders have the potential to provide complementary information across abnormalities. The predicted probabilities are averaged for each abnormality to reduce reliance on any single classifier.

\paragraph{Image measurements.} Report models may describe a finding while omitting its quantitative details. We therefore derive image measurements from TotalSegmentator v2 masks \citep{wasserthal2023totalsegmentator} and CT intensities in Hounsfield units (HU). These measurements supply numerical values and finding attributes for local revisions. Segmentation errors may introduce noise into these measurements. Their diagnostic meaning also depends on patient-specific characteristics, so a measurement alone does not always justify a diagnosis. We therefore learn rules to guide their use in report revision.

\subsection{Learning the Rule Bank}
\label{sec:selection}

For each report model, an agent learns a rule bank from revision outcomes across development cases. Each rule specifies a local edit, draft eligibility, and an optional condition on independent evidence. Learning starts with an empty rule bank. The agent receives descriptions and units of the available evidence, together with the target findings and available edits. In each iteration, the LLM proposes candidate rules by selecting a target finding, an edit, an evidence value, and a comparison direction. Code searches evidence thresholds on learning cases and applies each candidate rule to drafts satisfying its conditions. Candidate outcomes are evaluated by comparing revised reports with their original drafts against the reference reports. Feedback summarizes how many drafts each candidate changes and its effects on abnormality detection and report quality.

During learning, candidate rules are assessed using three criteria: (1) sufficient support across learning cases; (2) report quality assessed by GREEN; and (3) correctness of revisions to abnormality statements, assessed through CE-F1 or agreement with reference reports. After each iteration, the bank is updated using all candidates evaluated so far. It retains the best rule for each combination of target finding and edit type. The retained rules and candidate outcomes guide the agent's proposals in the next iteration. The agent then reviews the medical plausibility of the remaining evidence--finding relationships and removes unsupported rules. The final rule bank $\mathcal{B}_G$ is learned using all development cases for the corresponding report model.

\subsection{Report Revision}
\label{sec:revision}

During inference, an editor agent uses the fixed rule bank and independent evidence to revise each draft. Code identifies applicable rules by checking their conditions against the draft and independent evidence. The LLM follows the applicable rules to propose local edits, using any evidence values they require. Code checks and applies these edits while preserving the remaining report content. The editor agent may abstain and leave the draft unchanged. Each change is linked to its triggering rule, and the revision trace records any evidence values used.

\section{Experiments}
\label{sec:experiments}

\subsection{Experimental Setup}
\label{sec:setup}

\paragraph{Datasets.} CT-RATE pairs noncontrast 3D chest CT scans with radiology reports \citep{hamamci2026generalist}. It also provides labels for 18 abnormalities to support abnormality-level evaluation. The test set contains 1,564 scans from 1,304 patients in the official validation split. Report models produce more accurate drafts on scans seen during training, leaving fewer errors for rule learning. We therefore reserve 3,000 scans from 2,490 patients in the CT-RATE training split for development and exclude them from SelfCue-KD training.

\paragraph{Report Models.} The two report models are SelfCue-KD \citep{liang2026selfcue} and NV-Reason-CT \citep{myronenko2026nv}. SelfCue-KD distills contrastive decoding into model weights to better report findings already represented in the model. NV-Reason-CT is a SOTA 3D CT report generation model with strong abnormality-detection performance on CT-RATE. It provides a demanding test of whether our method could further improve an already strong report model.

\paragraph{Evaluation Metrics.} Existing metrics vary in their ability to reflect clinical errors \citep{xu2025radeval}. We therefore report multiple complementary metrics to assess report quality and textual similarity. Abnormality-detection performance is measured using CE-F1 and CRG \citep{hamamci2025crg}. CE-F1 denotes the macro-averaged F1 score over the 18 CT-RATE abnormalities extracted from reports by RadBERT \citep{yan2022radbert}. GREEN \citep{ostmeier2024green} assesses clinical errors against the reference report, while CT-FineBench \citep{yuan2026ct} evaluates answers to questions about findings and their attributes. Textual overlap is evaluated using BLEU-1 and BLEU-4 \citep{papineni2002bleu}, ROUGE-L \citep{lin2004rouge}, and METEOR \citep{banerjee2005meteor}.

\paragraph{Development Protocol.} Measurement definitions, edit operations, and addition wording were developed through exploratory analyses that included the public CT-RATE validation split used here as the test set. For the reported evaluation, evidence thresholds and rule-admission decisions are determined using development cases. Analyses on development cases informed the default precision requirement and GREEN margin for additions. The test results therefore reflect evaluation on a public benchmark that was inspected during method development.

\paragraph{Implementation Details.} For NV-Reason-CT, development drafts use a 3$^\circ$ input rotation to obtain an error level closer to the test setting. Five patient-level folds are used for development evaluation, with rules learned on four folds and evaluated on the held-out fold. Final rule banks are learned on all development cases before test evaluation. The learning agent and primary editor agent use gpt-oss-120b. The same model performs the medical plausibility review. The default precision requirement and GREEN margin for additions are fixed using development results. Further settings are given in Appendix~\ref{app:implementation-settings}. We use Qwen3-32B as the CT-FineBench judge, which yields scores similar to those of Qwen3-Max in the original study \citep{yuan2026ct}. For GREEN evaluation of NV-Reason-CT, reference sentences retain their original wording and are reorganized under the model's report section headings to reduce the influence of formatting differences. We estimate 95\% confidence intervals (CIs) using 2,000 paired bootstrap resamples at the patient level, taking the 2.5th and 97.5th percentiles. Experiments are run on NVIDIA B200 GPUs.

\subsection{Main Results}
\label{sec:results}

We evaluate rule bank revision on SelfCue-KD and NV-Reason-CT by comparing revised reports with their original drafts (Table~\ref{tab:main-performance}). For SelfCue-KD, rule bank revision increases CE-F1 by 11.5 points (CI $10.48$--$12.57$) and also improves CRG. To the best of our knowledge, NV-Reason-CT provides the strongest baseline for abnormality detection in 3D CT report generation. Even on this strong baseline, rule bank revision increases CE-F1 by 0.9 points (CI $0.49$--$1.29$) and also improves CRG. The detection breakdown shows higher macro recall for both models with relatively small changes in macro precision. The rule bank revises 85.0\% of SelfCue-KD drafts and 25.9\% of NV-Reason-CT drafts. On the changed NV-Reason-CT reports, CE-F1 increases by 2.3 points. These results support the value of learned rules for recovering omitted abnormality descriptions, even when the draft is produced by a strong report model.

For the changed SelfCue-KD reports, GREEN identifies fewer missing findings after revision (Table~\ref{tab:changed-reports}). SelfCue-KD gains 15.2 points overall on CT-FineBench, and the changed-report analysis shows improvements in both finding existence and attributes. These benefits are accompanied by more false findings in SelfCue-KD, helping explain the small decrease in its overall GREEN score. Overall GREEN changes remain modest, with CIs of $-1.38$ to $-0.03$ for SelfCue-KD and $-0.22$ to $0.57$ for NV-Reason-CT. SelfCue-KD also improves in BLEU-1, BLEU-4, and METEOR, with a slight decline in ROUGE-L, while NV-Reason-CT changes little across the four NLP metrics. The reference-based evaluations support targeted revision as a way to recover missing content and improve the reporting of finding attributes in SelfCue-KD drafts.

\begin{table}[t]
\caption{Performance on changed reports: 1,330 (85.0\%) for SelfCue-KD and 405 (25.9\%) for NV-Reason-CT. See Appendix~\ref{app:metric-components} for the full metric breakdown.}
\label{tab:changed-reports}
\centering\footnotesize
\definecolor{deltaBlueBackground}{RGB}{198,226,250}
\renewcommand{\markeddelta}[2]{\tikz[baseline=(delta.base)]{\node[fill=#1, text=black, rounded corners=1pt, inner xsep=0.9pt, inner ysep=0.5pt, outer sep=0pt] (delta) {#2};}}
\setlength{\tabcolsep}{1.5pt}
\renewcommand{\arraystretch}{1.08}
\begin{tabular}{@{}p{0.34\linewidth}rrr@{\hspace{8pt}}rrr@{}}
\toprule
Metric / statistic & \multicolumn{3}{c}{SelfCue-KD} & \multicolumn{3}{c}{NV-Reason-CT} \\
\cmidrule(lr){2-4}\cmidrule(lr){5-7}
 & Draft & Revised & $\Delta$ & Draft & Revised & $\Delta$ \\
\midrule
Changed (\%) & \multicolumn{3}{c}{85.0} & \multicolumn{3}{c}{25.9} \\
\midrule
\multicolumn{7}{l}{\textbf{(a) Abnormality detection}} \\
Macro precision $\uparrow$ & 57.5\phantom{0} & 58.4\phantom{0} & \markeddelta{deltaBlueBackground}{+0.8\phantom{0}} & 65.1\phantom{0} & 64.4\phantom{0} & \markeddelta{red!12}{-0.7\phantom{0}} \\
Macro recall $\uparrow$ & 43.4\phantom{0} & 60.1\phantom{0} & \markeddelta{deltaBlueBackground}{+16.7\phantom{0}} & 59.7\phantom{0} & 64.5\phantom{0} & \markeddelta{deltaBlueBackground}{+4.8\phantom{0}} \\
CE-F1 $\uparrow$ & 46.0\phantom{0} & 58.1\phantom{0} & \markeddelta{deltaBlueBackground}{+12.1\phantom{0}} & 61.7\phantom{0} & 64.0\phantom{0} & \markeddelta{deltaBlueBackground}{+2.3\phantom{0}} \\
\midrule
\multicolumn{7}{l}{\textbf{(b) GREEN}} \\
GREEN score $\uparrow$ & 48.5\phantom{0} & 47.7\phantom{0} & \markeddelta{red!12}{-0.8\phantom{0}} & 38.7\phantom{0} & 39.3\phantom{0} & \markeddelta{deltaBlueBackground}{+0.5\phantom{0}} \\
False findings $\downarrow$ & 1.22 & 1.71 & \markeddelta{red!12}{+0.49} & 1.60 & 1.63 & \markeddelta{red!12}{+0.04} \\
Missing findings $\downarrow$ & 3.32 & 3.01 & \markeddelta{deltaBlueBackground}{-0.31} & 3.20 & 3.14 & \markeddelta{deltaBlueBackground}{-0.06} \\
\midrule
\multicolumn{7}{l}{\textbf{(c) CT-FineBench}} \\
Overall CT-FineBench $\uparrow$ & 24.2\phantom{0} & 41.6\phantom{0} & \markeddelta{deltaBlueBackground}{+17.4\phantom{0}} & 45.6\phantom{0} & 44.8\phantom{0} & \markeddelta{red!12}{-0.8\phantom{0}} \\
Finding existence $\uparrow$ & 35.1\phantom{0} & 61.6\phantom{0} & \markeddelta{deltaBlueBackground}{+26.5\phantom{0}} & 68.6\phantom{0} & 67.9\phantom{0} & \markeddelta{red!12}{-0.7\phantom{0}} \\
All attributes $\uparrow$ & 21.1\phantom{0} & 35.5\phantom{0} & \markeddelta{deltaBlueBackground}{+14.4\phantom{0}} & 38.6\phantom{0} & 37.8\phantom{0} & \markeddelta{red!12}{-0.8\phantom{0}} \\
\bottomrule
\end{tabular}
\end{table}

\subsection{Role of Independent Evidence}
\label{sec:evidence-role}

We examine whether an agent could learn effective revision rules from draft-derived signals alone and how independent image evidence affects this learning process (Table~\ref{tab:independent-evidence}). Draft-only learning uses report-derived signals under the same revision-outcome feedback, admission criteria, and medical plausibility review as the combined-evidence setting but has no independent observation of the scan. The agent learns eight draft-only rules for SelfCue-KD and five for NV-Reason-CT before post-learning checks. These checks retain only one and two rules, respectively. Most rules are removed because the learning agent finds insufficient medical support for the relationship between their draft-derived signals and target abnormalities. The retained banks revise only a small fraction of reports and produce little change in CE-F1. Even without medical plausibility review, CE-F1 gains remain limited to 0.2 points for SelfCue-KD and 0.3 points for NV-Reason-CT. Appendix~\ref{app:draft-only-rules} presents draft-only rule examples and analyzes their limited revision gains. These results highlight the need for independent image evidence, as draft-only learning provides insufficient information for learning effective revision rules.

Although classifier predictions contain noise, rules learned from these predictions improve CE-F1 for both report models. Image measurements also support abnormality detection within the 18 CT-RATE labels and provide information about abnormalities outside this label set. For example, rules using liver attenuation improve reference-based F1 for hepatic steatosis from 34.5 to 56.4 for SelfCue-KD and from 0.0 to 51.0 for NV-Reason-CT. Hepatic-steatosis labels are extracted from reference reports, drafts, and revised reports using the same gpt-oss-120b yes/no prompt. F1 is computed against the labels extracted from reference reports. Combining classifier predictions and image measurements yields the highest CE-F1 and CRG, suggesting that the two evidence sources provide complementary information for report revision.

\begin{table}[t]
\caption{Rule learning with draft-derived signals and different independent evidence sources. Rule banks are learned separately for each evidence setting.}
\label{tab:independent-evidence}
\centering\scriptsize
\setlength{\tabcolsep}{3pt}
\renewcommand{\arraystretch}{1.12}
\begin{tabular}{@{}p{0.40\linewidth}rrrr@{}}
\toprule
Setting & CE-F1 & CRG & GREEN & CT-FineBench \\
\midrule
\multicolumn{5}{l}{\textbf{SelfCue-KD}} \\
Draft & 46.2 & 43.9 & 50.5 & 26.2 \\
Draft-only learning & 46.2 & 43.9 & 50.4 & 26.3 \\
Classifier evidence & 54.5 & 50.8 & 50.1 & 39.6 \\
Measurement evidence & 50.9 & 45.4 & 50.6 & 28.9 \\
Both evidence sources & 57.7 & 51.9 & 49.8 & 41.4 \\
\addlinespace[4pt]
\multicolumn{5}{l}{\textbf{NV-Reason-CT}} \\
Draft & 63.0 & 54.0 & 45.8 & 44.9 \\
Draft-only learning & 63.1 & 54.1 & 45.8 & 44.9 \\
Classifier evidence & 63.8 & 54.8 & 45.9 & 44.6 \\
Measurement evidence & 63.4 & 54.3 & 45.7 & 45.0 \\
Both evidence sources & 63.9 & 54.9 & 45.9 & 44.7 \\
\bottomrule
\end{tabular}
\end{table}

\subsection{Ablation Studies and Analysis}
\label{sec:ablation}

\paragraph{Report Revision Strategies.} We compare direct LLM revision and fixed decision strategies with the learned rule bank to assess the value of controlled use of independent evidence (Table~\ref{tab:direct-revision}). The direct LLM control is inspired by MedScribe's final synthesis stage \citep{orlando2026medscribe} and uses the same editor agent and evidence sources as rule bank revision. Direct LLM revision and classifier-driven revision improve abnormality detection for SelfCue-KD, with classifier-driven revision achieving the largest gain. However, both strategies lower GREEN for SelfCue-KD and reduce both detection performance and GREEN for NV-Reason-CT. The fixed clinical rubric, based on literature-informed measurement cutoffs, also reduces detection performance for both report models, whereas learning rules from measurement evidence improves it. One limitation of fixed cutoffs is that normal anatomical dimensions vary with patient characteristics such as age, sex, and body size \citep{wolak2008aortic,doi:10.1161/CIR.0000000000001106}. The learned rule bank instead improves detection for both report models while keeping GREEN closer to their draft levels than the alternative strategies. These comparisons show that evidence sources alone do not determine revision benefits: the decisions governing their use must account for the report model's existing strengths and errors.

\begin{table}[t]
\caption{Comparison of report revision strategies using the same drafts. Direct LLM revision lets the LLM select edits from the draft and independent evidence. Classifier-driven revision uses development-calibrated classifier thresholds. The fixed clinical rubric uses literature-informed measurement cutoffs to guide report revision. Rule Bank uses learned conditions on the draft and evidence to select local edits.}
\label{tab:direct-revision}
\centering\scriptsize
\setlength{\tabcolsep}{3pt}
\renewcommand{\arraystretch}{1.12}
\begin{tabular}{@{}p{0.36\linewidth}rrrr@{}}
\toprule
Revision strategy & CE-F1 & CRG & GREEN & \shortstack{CT-\\FineBench} \\
\midrule
\multicolumn{5}{l}{\textbf{SelfCue-KD}} \\
Draft & 46.2 & 43.9 & 50.5 & 26.2 \\
Direct LLM revision & 52.5 & 49.1 & 43.8 & 42.4 \\
Classifier-driven & 58.9 & 53.7 & 48.9 & 44.0 \\
Fixed clinical rubric & 44.4 & 43.5 & 49.3 & 28.3 \\
Rule Bank & 57.7 & 51.9 & 49.8 & 41.4 \\
\addlinespace[3pt]
\multicolumn{5}{l}{\textbf{NV-Reason-CT}} \\
Draft & 63.0 & 54.0 & 45.8 & 44.9 \\
Direct LLM revision & 57.1 & 54.1 & 39.2 & 51.4 \\
Classifier-driven & 60.1 & 54.8 & 42.9 & 45.5 \\
Fixed clinical rubric & 61.2 & 53.1 & 45.0 & 46.4 \\
Rule Bank & 63.9 & 54.9 & 45.9 & 44.7 \\
\bottomrule
\end{tabular}
\end{table}

\begin{figure}[t]
\centering
\includegraphics[width=\linewidth]{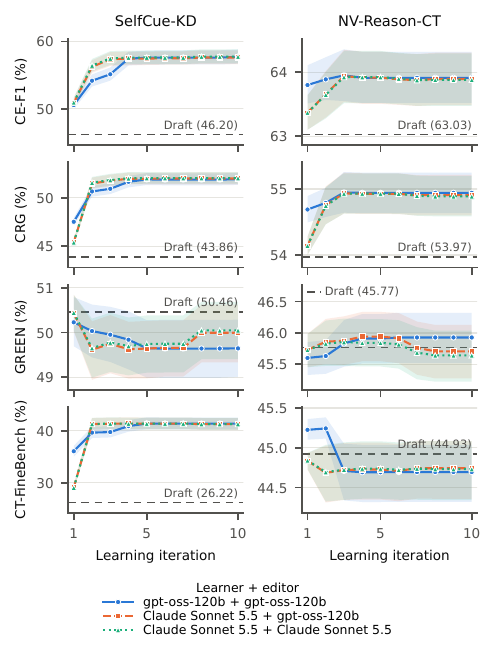}
\caption{Revision performance over ten learning iterations on the test set. Dashed lines show draft scores. Shading shows 95\% CIs for changes from the draft, shifted by the draft scores.}
\label{fig:learning-iterations}
\end{figure}

\paragraph{Iterative Rule Learning.} We examine how revision performance evolves across learning iterations and whether this trend is consistent across different combinations of learning agents and editor agents (Figure~\ref{fig:learning-iterations}). Each iteration's rule bank is evaluated on the same test scans. Most abnormality-detection gains emerge within the first three to four iterations for both report models, followed by smaller changes. Despite differences in their early trajectories, the tested agent combinations reach similar abnormality-detection performance. With the Sonnet learning agent held fixed, changing the editor agent also has little effect on this performance. For SelfCue-KD, improvements in detection and finding-level question answering are accompanied by a small decrease in GREEN. Overall, the learning process reaches stable detection performance within a few iterations across the tested agent implementations.

\paragraph{Effect of the Precision Requirement.} We vary the minimum precision required for accepting rules that add labeled abnormalities to examine its effect on revision coverage and performance (Table~\ref{tab:precision-threshold}). Precision is measured against development reference labels, and a separate rule bank is learned for each threshold. For SelfCue-KD, raising the threshold from 50\% to 70\% reduces the proportion of reports revised from 85.0\% to 49.1\%, with the CE-F1 gain decreasing from 11.5 to 6.0 points. At the same time, overall GREEN moves closer to the draft score. NV-Reason-CT shows the same broad trade-off, although the changes across thresholds are not strictly monotonic. Lower precision requirements generally allow broader revision and larger detection gains, accompanied by lower GREEN scores. The precision requirement therefore provides a practical way to adjust how broadly the system revises reports while balancing detection gains and reference-based report quality.

\begin{table}[t]
\caption{Effect of the precision requirement on revision. $\Delta$ denotes overall changes from the draft (points).}
\label{tab:precision-threshold}
\centering\scriptsize
\setlength{\tabcolsep}{3pt}
\renewcommand{\arraystretch}{1.12}
\begin{tabular}{@{}lrrr@{\hspace{8pt}}rrr@{}}
\toprule
 & \multicolumn{3}{c}{SelfCue-KD} & \multicolumn{3}{c}{NV-Reason-CT} \\
\cmidrule(lr){2-4}\cmidrule(lr){5-7}
\shortstack{Min. precision\\(\%)} & \shortstack{Changed\\(\%)} & $\Delta$CE-F1 & $\Delta$GREEN & \shortstack{Changed\\(\%)} & $\Delta$CE-F1 & $\Delta$GREEN \\
\midrule
30 & 86.0 & +12.21 & -1.69 & 51.3 & +1.07 & -0.65 \\
40 & 86.2 & +11.79 & -1.18 & 42.5 & +1.14 & -0.26 \\
50 & 85.0 & +11.50 & -0.69 & 25.9 & +0.89 & +0.17 \\
60 & 60.1 & +8.81 & 0.00 & 19.9 & +0.55 & +0.11 \\
70 & 49.1 & +6.02 & +0.08 & 20.0 & +0.42 & +0.11 \\
\bottomrule
\end{tabular}
\end{table}

%

\paragraph{Robustness to Draft Variation.} We assess whether revision gains persist across different sampled drafts using five seeds (Table~\ref{tab:draft-variation}). For each seed, we sample development and test drafts and learn a separate rule bank from the development drafts. The report-model checkpoint, independent evidence, learning procedure, and editor agent configuration remain fixed. CE-F1 improves for both report models across all five seeds. GREEN remains close to the draft score on average.

\begin{table}[t]
\caption{Revision coverage and performance changes across five sampling seeds. Values are mean $\pm$ sample SD. A separate rule bank is learned for each seed.}
\label{tab:draft-variation}
\centering\scriptsize
\setlength{\tabcolsep}{4pt}
\renewcommand{\arraystretch}{1.12}
\begin{tabular}{@{}lrrr@{}}
\toprule
Report model & Changed (\%) & $\Delta$CE-F1 & $\Delta$GREEN \\
\midrule
SelfCue-KD & $83.4\pm 1.9$ & $+10.67\pm 0.80$ & $-0.06\pm 0.15$ \\
NV-Reason-CT & $24.0\pm 2.9$ & $+0.72\pm 0.33$ & $+0.10\pm 0.26$ \\
\bottomrule
\end{tabular}
\end{table}

\paragraph{Quantitative Information.} We examine how revision supplements drafts with missing quantitative information (Table~\ref{tab:quantitative-information}). Numerical details are sparse in the original drafts, particularly those from NV-Reason-CT. Revision adds 558 vascular diameter values across both report models, all matching the supplied measurements after rounding. On the matched ascending-aorta subset, the added values differ from reference measurements by 2.3--2.4 mm on average. For context, a dedicated automatic measurement method reported a mean absolute diameter error of 1.09 mm against manual annotations at predefined levels \citep{sedghi2019automated}. The revision trace records the rule and source measurement behind each added value, providing a clear starting point for clinicians to inspect the underlying segmentation and assess measurement reliability. Overall, these results illustrate how independent image measurements enrich reports with quantitative anatomical details missing from the original drafts.

\begin{table}[t]
\caption{Quantitative reporting and reference agreement. MAE is computed on added ascending-aorta diameters with matching reference measurements.}
\label{tab:quantitative-information}
\centering\scriptsize
\setlength{\tabcolsep}{4pt}
\renewcommand{\arraystretch}{1.12}
\begin{tabular}{@{}lrrr@{}}
\toprule
\multicolumn{4}{l}{\textbf{Quantitative reporting on the test set}} \\
\addlinespace[2pt]
Report model & Draft (\%) & Revised (\%) & Added measurements \\
\midrule
SelfCue-KD & 18.9 & 31.3 & 517 \\
NV-Reason-CT & 0.3 & 2.9 & 41 \\
\midrule
\multicolumn{4}{l}{\textbf{Ascending aorta: agreement with reference reports}} \\
\addlinespace[2pt]
Report model & Added & Matched & MAE (mm) \\
\midrule
SelfCue-KD & 240 & 62 & 2.3 \\
NV-Reason-CT & 41 & 23 & 2.4 \\
\bottomrule
\end{tabular}
\end{table}

\subsection{Quality Analysis}
\label{sec:case-studies}

\begin{figure*}[t]
\centering
\includegraphics[width=\textwidth]{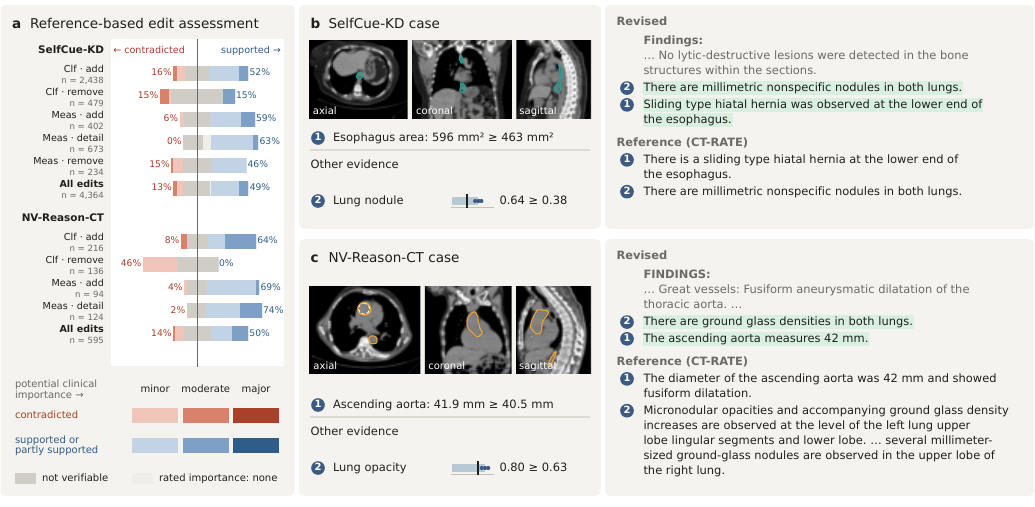}
\caption{Edit quality assessment and illustrative cases. (a) Reference-report agreement and potential clinical importance of revisions. (b,c) Examples linking independent evidence to local report edits.}
\label{fig:quality-cases}
\end{figure*}

We assess individual revisions to examine both their correctness and potential clinical value (Figure~\ref{fig:quality-cases}). Claude Sonnet 5.5 serves as an LLM-based radiology reviewer to assess each edit against the reference report. Supported or partly supported edits outnumber contradictions for both report models. Measurement-based additions are less often contradicted than classifier-based additions. For measurement-based detail edits, contradiction rates are 0.3\% for SelfCue-KD and 2.4\% for NV-Reason-CT. Classifier-based removals require greater caution: 45.6\% contradict the reference for NV-Reason-CT. The assessment supports using independent evidence to enrich reports with omitted findings and quantitative details through controlled revisions. Partial support may still involve incorrect details, while reference silence leaves edits unverifiable. Assessment criteria and potential clinical importance ratings are detailed in Appendix~\ref{app:case-analysis}.

The two cases illustrate how independent evidence adds useful information to drafts. For SelfCue-KD, an esophagus measurement supports adding hiatal hernia and classifier evidence supports adding lung nodules. The displayed segmentation and classifier scores show how different evidence sources support distinct additions within the same report. Each evidence value satisfies a learned rule condition that specifies when the corresponding addition is warranted. For NV-Reason-CT, the draft identifies aortic dilation but omits its diameter. The measured diameter of 41.9 mm is rounded to 42 mm in the revision, matching the reference. Classifier evidence also supports adding lung opacity. All illustrated additions are supported by the reference reports. The links between evidence and revised text make the basis for each addition explicit, providing clinicians with a concrete starting point for reviewing the changes and their supporting measurements or predictions.

\section{Discussion and Limitations}
\label{sec:discussion}

\noindent\textbf{Complementing Report Models.} The rule bank complements report models by using independent image evidence to guide local revisions. Gains are larger for SelfCue-KD, while the stronger NV-Reason-CT also benefits. These improvements require no retraining of the report model. We expect this approach to be particularly useful when a model encounters unfamiliar data and needs additional observations to support report revision.

\noindent\textbf{Model Learning from Rules.} The rule bank forms part of the agent harness, which organizes the rules and tools supporting the report model. It records knowledge and reveals weaknesses such as hepatic-steatosis omissions in NV-Reason-CT. We envision two roles for this knowledge. Reusable recognition and reporting knowledge could guide model training through verified revisions. Knowledge about obtaining and interpreting scan-specific measurements is better retained in the harness to guide tool calls when needed. Future work will connect these roles in a self-evolving agent system: agent experience updates the rule bank, selected knowledge guides model training, and outcomes from the updated model support rule learning \citep{lee2026recursive}.

\noindent\textbf{Limitation: Evidence Quality.} We screen image measurements and learn rules to control how noisy evidence informs revision. Yet a fixed clinical rubric may leave borderline findings ambiguous even when measurements are accurate. Anatomical masks also provide limited lesion-specific information. The agent receives evidence values rather than images and cannot visually reassess such cases. Expanding measurement coverage and incorporating targeted image review are future directions.

\noindent\textbf{Limitation: Clinical Validation.} Our quality analysis uses an LLM as a radiology reviewer with prompts developed under radiologist guidance. This guidance brings clinical expertise into the assessment criteria, although the evaluations themselves are performed by the LLM. Its interpretation of individual revisions may therefore differ from that of a radiologist, which should be considered when interpreting these assessments.

\section{Conclusion}
\label{sec:conclusion}

We presented an agent system that learns model-specific executable rule banks for revising 3D CT reports with independent image evidence. Revision outcomes across development cases guide when classifier predictions and image measurements warrant local edits. On CT-RATE, the banks improve abnormality detection for SelfCue-KD and NV-Reason-CT across draft samples. The limited gains from draft-only learning highlight the value of additional image observations for learning effective revision rules. Each revision is traceable to its triggering rule and any evidence used. Future work will explore transferring reusable knowledge into report models through training on verified revisions while retaining measurement tools in the harness. Revision outcomes from updated models could then guide further rule learning toward a self-evolving agent system.

\clearpage
\section*{LLM Usage Statement}
\label{sec:llm-usage-statement}

OpenAI Codex and Anthropic Claude assisted with manuscript drafting and editing, literature discovery, discussion of paper framing and result interpretation, code development, figure preparation, and internal consistency checks. Experimental uses of gpt-oss-120b, Claude Sonnet 5.5, GREEN-RadLlama2-7b, and Qwen3-32B include rule learning, report revision, reference-based edit assessment, and metric evaluation, as detailed in the main text and appendix. The authors retain responsibility for the manuscript and all scientific claims.

{
    \small
    \bibliography{google_scholar}
}

\clearpage
\appendix

\begin{table*}[!b]
\caption{Image measurements and coverage across our training, development, and test sets in CT-RATE.}
\label{tab:measurements}
\centering\scriptsize
\setlength{\tabcolsep}{3pt}
\renewcommand{\arraystretch}{1.18}
\begin{tabular}{@{}p{0.16\textwidth}p{0.30\textwidth}p{0.12\textwidth}p{0.29\textwidth}r@{}}
\toprule
Measurement & Definition & Source regions & Processing and checks & \shortstack[r]{Coverage\\(\%)} \\
\midrule
Cardiothoracic ratio & Largest left-right heart width over the axial slices divided by the largest inner width of the rib cage on the slices that contain the heart & heart, ribs & -- & 99.97 \\
Coronary calcium (mm$^3$) & Volume of voxels of at least 130 HU in the coronary region & coronary arteries, bones & Dilate coronary mask by 2 mm; discard components below 1 mm$^3$; exclude bone within 2 mm and metal. Use a soft reconstruction or smooth a sharp-only scan ($\sigma=1$ mm). & 99.99 \\
Aortic wall calcium (mm$^3$) & Volume of voxels of at least 130 HU in a shell extending 2 mm inside and outside the aortic boundary & aorta & Apply the same bone, metal, and small-component exclusions; exclude the coronary region and slices flagged for mask leakage. Use the same reconstruction handling as for coronary calcium. & 99.99 \\
Pericardial fluid (mL) & Sum of pericardial-effusion component volumes & pericardial effusion & Assign 0 mL when no fluid component is recorded. & 100.00 \\
Pleural fluid (mL) & Total pleural-effusion component volume across both sides & pleural effusion & Assign 0 mL when no fluid component is recorded. & 100.00 \\
Distal esophagus area (mm$^2$) & Largest axial cross-section of the esophagus mask within its lowest 60 mm & esophagus & Locate the lower boundary at the lowest slice with at least 20 mm$^2$ of esophagus. & 99.99 \\
Internal metal (mm$^3$) & Volume of voxels of at least 3000 HU classified as internal metal & thoracic organs, vessels, bones & Require a component to lie within 3 mm of a thoracic structure and at least 5 mm below the body-mask boundary. & 100.00 \\
Ascending aorta diameter (mm) & Largest inscribed diameter of cross-sections perpendicular to the aortic centreline, ascending segment & aorta & Exclude cross-sections touching the sampling-window edge; unavailable segments remain missing. & 99.16 \\
Aortic arch diameter (mm) & Largest inscribed diameter of cross-sections perpendicular to the aortic centreline, arch segment & aorta & Exclude cross-sections touching the sampling-window edge; unavailable segments remain missing. & 99.15 \\
Descending aorta diameter (mm) & Largest inscribed diameter of cross-sections perpendicular to the aortic centreline, descending thoracic segment & aorta & Exclude cross-sections touching the sampling-window edge; unavailable segments remain missing. & 99.16 \\
Main pulmonary artery diameter (mm) & Largest inscribed diameter of the largest pulmonary-artery component over axial slices & pulmonary artery & Apply the fixed $-3.8$ mm development calibration; an absent pulmonary-artery mask gives a missing value. & 99.95 \\
Liver attenuation (HU) & Median CT attenuation within the liver region & liver & Erode the mask by 2 voxels and require at least 50 remaining voxels. Supply the negative median HU as evidence. & 99.95 \\
Spine coronal angle range ($^\circ$) & Range of coronal-plane angles between consecutive vertebral centroids & vertebrae & Use vertebrae with at least 300 voxels; require at least five vertebrae. Smooth angles over three vertebrae before taking their range. & 99.99 \\
Pleural effusion side & Left, right, or bilateral pleural effusion, determined from component centroids & pleural effusion & Ignore components within 25 voxels of the midline; require at least 10 mL on a side to mark it present. & 100.00 \\
Device type & Device category inferred from internal metal: sternotomy wires, pacemaker / ICD, valve prosthesis, or spinal fixation & thoracic organs, vessels, bones & Apply rules on anatomical location, component volume, skin depth, elongation, and overlap with source regions. & 100.00 \\
\bottomrule
\end{tabular}
\end{table*}

\section{Additional Method Details}
\label{app:method-details}

\subsection{Image Measurements}
\label{app:image-measurements}
 Image measurements are derived from TotalSegmentator v2 masks \citep{wasserthal2023totalsegmentator} and CT intensities (Table~\ref{tab:measurements}). They include numerical quantities and categorical details such as effusion side and device type. Coverage is measured over 25,692 scans: 21,128 training, 3,000 development, and 1,564 test scans. A measured volume of zero counts as an available measurement. Coverage for side and device type follows the availability of the underlying effusion and metal measurements. Coverage indicates measurement availability rather than segmentation or diagnostic accuracy.

Calcium smoothing was selected using 100 development scans with paired soft and sharp reconstructions. Liver attenuation uses unsmoothed CT intensities. Aortic centrelines are extracted at 1 mm isotropic resolution, with perpendicular cross-sections measured on the original masks. Pulmonary-artery diameters use a fixed $-3.8$ mm offset fitted to development reference measurements. This calibration is applied to development and test measurements.

Measurements with documented extraction failures are excluded. Missing values do not satisfy evidence conditions or support diameter edits. Flags from aortic segmentation quality checks are recorded for inspection and do not filter the evidence. Author review informed measurement choices and selected quality checks without case-by-case mask correction.

\subsection{Rule Admission}
\label{app:rule-admission}

Code admits rules that change at least 20 of 3,000 development drafts, scaling this minimum to the number of learning cases. Candidate rules must satisfy GREEN-based report-quality checks. Label-based additions and removals must improve CE-F1, with at least 50\% precision against reference labels additionally required for additions. Outside the label set, the target abnormality must appear in at least 60\% of affected reference reports for additions and at most 40\% for removals. Detail edits use only the support and GREEN checks. For each target and edit, the bank retains the eligible rule with the largest CE-F1 gain for label-based edits and the largest number of changed drafts otherwise.

After learning, the learning agent reviews medical plausibility for rules with evidence conditions. Edits that directly use measured effusion side or device type are exempt. The review uses only the target, edit, evidence description, and comparison direction. Unsupported relationships are rejected, and the same review decision is reused across folds. This assesses the relationship rather than the numerical cutoff or every resulting revision.

\subsection{Report Revision}
\label{app:report-revision-details}

Local edits add or remove findings, supply missing measurements, and update finding attributes. For the 18 labeled abnormalities, eligibility uses RadBERT draft labels; other targets are checked directly in the draft text. Numerical and categorical edits require the corresponding image measurement. During learning, code simulates these operations to evaluate candidate rules and may abstain when an edit cannot preserve other findings in the same sentence. At inference, the editor agent follows the same operation constraints and is instructed to preserve findings outside the rule's target. It receives numbered draft sentences, relevant evidence values, and one instruction per applicable rule, then returns rule-linked edits that rewrite or remove numbered sentences or add sentences. Code checks rule and sentence references before applying edits, keeps repeated output sentences once, and copies unedited sentences unchanged. An empty edit list or failure to return applicable edits after the configured retries leaves the original draft unchanged. Device descriptions use sternotomy wires, pacemakers or implantable cardioverter-defibrillators (ICDs), valve prostheses, and spinal fixation.

\section{Experimental and Evaluation Details}
\label{app:experimental-details}

\subsection{Implementation Settings}
\label{app:implementation-settings}

\paragraph{Classifier evidence.} Linear classification heads are trained on fixed encoder features. Development scores are generated using five patient-level folds of the CT-RATE training partition. Each fold is scored by heads trained on the other four. Rule learning uses a separate patient-level split and reuses these precomputed scores. For test prediction, the final classifiers use 21,765 training scans for fitting and 2,363 training scans for early stopping.

\paragraph{Learning and sampling.} The default learning agent uses gpt-oss-120b for rule proposals and medical plausibility checks. For the draft-variation experiments, each report model samples development and test drafts using five random seeds with temperature 0.7 and top-$p$ 0.9. A separate bank is learned from the development drafts for each seed. Each learning run uses 20 iterations with a budget of 20 new proposals per iteration. Code searches 11 percentile-based thresholds among eligible learning cases for each proposed condition. Each condition compares one evidence value with one threshold. Duplicate proposals are rejected. Rule admission requires the mean paired GREEN change on changed reports to be at least $-1.0$ points for additions of the 18 labeled abnormalities and nonnegative for other edits.

\paragraph{Editor agent.} The primary editor agent uses gpt-oss-120b at temperature 0 with an output-token limit of 16,384. A request may be retried twice at temperature 0.7 when its returned edit list cannot be applied. The Claude Sonnet 5.5 comparisons use the same rule instructions and response format through Claude Code, with sampling settings not controlled by the local configuration.

\subsection{Evaluation Details}
\label{app:evaluation-details}

CE-F1, CRG, and NLP metrics use all 1,564 test reports. GREEN uses reports with valid scores, while CT-FineBench uses reports with benchmark questions. Changed-report analyses recompute detection metrics on the modified subset and use paired valid evaluations for GREEN and CT-FineBench. Unchanged reports retain their original text and evaluation scores. NLP evaluation uses the original CT-RATE references, while GREEN references for NV-Reason-CT are reorganized to match its structured output. GREEN uses GREEN-RadLlama2-7b at temperature 0, with each report limited to 800 words and over-budget prompts truncated from the end. CT-FineBench uses Qwen3-32B with an answer-only prompt and compares its answers with reference answers \citep{yuan2026ct}.

\subsection{Development Preprocessing Checks}
\label{app:development-boundaries}

Development evaluation learns rules on four patient folds and evaluates them on the remaining fold. Threshold search and admission use only the learning folds. The check for evidence conditions that select nearly all eligible drafts uses evidence values from all development cases, including the held-out fold, without held-out reference labels. Restricting it to learning folds leaves CE-F1 gains unchanged at the reported precision and produces only small changes in GREEN and edit counts (Table~\ref{tab:development-boundaries}). The $-3.8$ mm pulmonary-artery calibration also uses all development reference reports, including held-out folds. Refitting it from learning folds changes one held-out diameter by 1 mm in the learned rule banks. This check holds the banks fixed and does not reevaluate GREEN-based rule admission after recalibration, so its full effect on rule selection remains unmeasured. These checks leave the final banks and test evaluation unchanged.

\begin{table}[t]
\caption{Development estimates before and after restricting condition simplification to learning folds.}
\label{tab:development-boundaries}
\centering\scriptsize
\setlength{\tabcolsep}{0.5pt}
\renewcommand{\arraystretch}{1.12}
\begin{tabular*}{0.96\linewidth}{@{\extracolsep{\fill}}lrrrr@{}}
\toprule
Report model & \shortstack{Changed\\decisions} & \shortstack{Held-out\\$\Delta$CE-F1} & \shortstack{Held-out\\$\Delta$GREEN} & Edits \\
\midrule
SelfCue-KD & 0 & $+11.55\to +11.55$ & $+0.19\to +0.19$ & $6,390\to 6,390$ \\
NV-Reason-CT & 2 & $+0.85\to +\phantom{0}0.85$ & $+0.46\to +0.47$ & $1,120\to 1,128$ \\
\bottomrule
\end{tabular*}
\end{table}

\section{Additional Main Results}
\label{app:additional-main-results}

\subsection{Detailed Performance Results}
\label{app:metric-components}

\begin{table}[t]
\caption{Full metric breakdown on changed reports. GREEN components are mean counts per report; attribute groups use reports with relevant questions.}
\label{tab:changed-reports-full}
\centering\scriptsize
\setlength{\tabcolsep}{1.5pt}
\renewcommand{\arraystretch}{1.08}
\begin{tabular}{@{}p{0.34\linewidth}rrr@{\hspace{8pt}}rrr@{}}
\toprule
Metric / statistic & \multicolumn{3}{c}{SelfCue-KD} & \multicolumn{3}{c}{NV-Reason-CT} \\
\cmidrule(lr){2-4}\cmidrule(lr){5-7}
 & Draft & Revised & $\Delta$ & Draft & Revised & $\Delta$ \\
\midrule
Changed (\%) & \multicolumn{3}{c}{85.0} & \multicolumn{3}{c}{25.9} \\
\midrule
\multicolumn{7}{l}{\textbf{(a) Abnormality detection} ($n=1{,}330/405$)} \\
Macro precision $\uparrow$ & 57.5\phantom{0} & 58.4\phantom{0} & \markeddelta{deltaBlueBackground}{+0.8\phantom{0}} & 65.1\phantom{0} & 64.4\phantom{0} & \markeddelta{red!12}{-0.7\phantom{0}} \\
Macro recall $\uparrow$ & 43.4\phantom{0} & 60.1\phantom{0} & \markeddelta{deltaBlueBackground}{+16.7\phantom{0}} & 59.7\phantom{0} & 64.5\phantom{0} & \markeddelta{deltaBlueBackground}{+4.8\phantom{0}} \\
CE-F1 $\uparrow$ & 46.0\phantom{0} & 58.1\phantom{0} & \markeddelta{deltaBlueBackground}{+12.1\phantom{0}} & 61.7\phantom{0} & 64.0\phantom{0} & \markeddelta{deltaBlueBackground}{+2.3\phantom{0}} \\
CRG $\uparrow$ & 43.3\phantom{0} & 51.7\phantom{0} & \markeddelta{deltaBlueBackground}{+8.3\phantom{0}} & 50.3\phantom{0} & 52.5\phantom{0} & \markeddelta{deltaBlueBackground}{+2.1\phantom{0}} \\
True positives (count) $\uparrow$ & 2,054 & 3,202 & \markeddelta{deltaBlueBackground}{+1,148} & 1,212 & 1,316 & \markeddelta{deltaBlueBackground}{+104} \\
False positives (count) $\downarrow$ & 1,326 & 2,199 & \markeddelta{red!12}{+873} & 616 & 684 & \markeddelta{red!12}{+68} \\
False negatives (count) $\downarrow$ & 2,887 & 1,739 & \markeddelta{deltaBlueBackground}{-1,148} & 736 & 632 & \markeddelta{deltaBlueBackground}{-104} \\
\midrule
\multicolumn{7}{l}{\textbf{(b) GREEN components} ($n=1{,}330/395$)} \\
GREEN score $\uparrow$ & 48.5\phantom{0} & 47.7\phantom{0} & \markeddelta{red!12}{-0.8\phantom{0}} & 38.7\phantom{0} & 39.3\phantom{0} & \markeddelta{deltaBlueBackground}{+0.5\phantom{0}} \\
Matched findings $\uparrow$ & 5.08 & 5.08 & \markeddelta{white}{+0.00} & 3.16 & 3.22 & \markeddelta{deltaBlueBackground}{+0.06} \\
Total significant errors $\downarrow$ & 5.03 & 5.26 & \markeddelta{red!12}{+0.23} & 5.27 & 5.30 & \markeddelta{red!12}{+0.03} \\
False findings $\downarrow$ & 1.22 & 1.71 & \markeddelta{red!12}{+0.49} & 1.60 & 1.63 & \markeddelta{red!12}{+0.04} \\
Missing findings $\downarrow$ & 3.32 & 3.01 & \markeddelta{deltaBlueBackground}{-0.31} & 3.20 & 3.14 & \markeddelta{deltaBlueBackground}{-0.06} \\
Incorrect location $\downarrow$ & 0.17 & 0.23 & \markeddelta{red!12}{+0.06} & 0.24 & 0.27 & \markeddelta{red!12}{+0.03} \\
Incorrect severity $\downarrow$ & 0.12 & 0.14 & \markeddelta{red!12}{+0.02} & 0.13 & 0.12 & \markeddelta{deltaBlueBackground}{-0.01} \\
Unsupported comparisons $\downarrow$ & 0.09 & 0.08 & \markeddelta{deltaBlueBackground}{-0.01} & 0.03 & 0.04 & \markeddelta{red!12}{+0.01} \\
Missing prior-study changes $\downarrow$ & 0.11 & 0.09 & \markeddelta{deltaBlueBackground}{-0.02} & 0.06 & 0.09 & \markeddelta{red!12}{+0.03} \\
\midrule
\multicolumn{7}{l}{\textbf{(c) CT-FineBench} ($n=1{,}209/393$)} \\
Overall CT-FineBench $\uparrow$ & 24.2\phantom{0} & 41.6\phantom{0} & \markeddelta{deltaBlueBackground}{+17.4\phantom{0}} & 45.6\phantom{0} & 44.8\phantom{0} & \markeddelta{red!12}{-0.8\phantom{0}} \\
Finding existence $\uparrow$ & 35.1\phantom{0} & 61.6\phantom{0} & \markeddelta{deltaBlueBackground}{+26.5\phantom{0}} & 68.6\phantom{0} & 67.9\phantom{0} & \markeddelta{red!12}{-0.7\phantom{0}} \\
All attributes $\uparrow$ & 21.1\phantom{0} & 35.5\phantom{0} & \markeddelta{deltaBlueBackground}{+14.4\phantom{0}} & 38.6\phantom{0} & 37.8\phantom{0} & \markeddelta{red!12}{-0.8\phantom{0}} \\
\quad Location $\uparrow$ & 23.3\phantom{0} & 30.4\phantom{0} & \markeddelta{deltaBlueBackground}{+7.1\phantom{0}} & 43.3\phantom{0} & 41.9\phantom{0} & \markeddelta{red!12}{-1.4\phantom{0}} \\
\quad Size $\uparrow$ & 9.7\phantom{0} & 42.9\phantom{0} & \markeddelta{deltaBlueBackground}{+33.2\phantom{0}} & 11.3\phantom{0} & 12.8\phantom{0} & \markeddelta{deltaBlueBackground}{+1.6\phantom{0}} \\
\quad Severity $\uparrow$ & 24.0\phantom{0} & 33.2\phantom{0} & \markeddelta{deltaBlueBackground}{+9.2\phantom{0}} & 40.9\phantom{0} & 40.4\phantom{0} & \markeddelta{red!12}{-0.5\phantom{0}} \\
\quad Morphology $\uparrow$ & 24.2\phantom{0} & 35.8\phantom{0} & \markeddelta{deltaBlueBackground}{+11.5\phantom{0}} & 37.1\phantom{0} & 39.6\phantom{0} & \markeddelta{deltaBlueBackground}{+2.5\phantom{0}} \\
\quad Distribution $\uparrow$ & 27.0\phantom{0} & 32.3\phantom{0} & \markeddelta{deltaBlueBackground}{+5.3\phantom{0}} & 41.1\phantom{0} & 38.4\phantom{0} & \markeddelta{red!12}{-2.7\phantom{0}} \\
\quad Density $\uparrow$ & 16.0\phantom{0} & 19.4\phantom{0} & \markeddelta{deltaBlueBackground}{+3.5\phantom{0}} & 24.3\phantom{0} & 27.2\phantom{0} & \markeddelta{deltaBlueBackground}{+2.8\phantom{0}} \\
\quad Other attributes $\uparrow$ & 23.1\phantom{0} & 38.4\phantom{0} & \markeddelta{deltaBlueBackground}{+15.3\phantom{0}} & 40.3\phantom{0} & 39.5\phantom{0} & \markeddelta{red!12}{-0.7\phantom{0}} \\
No-answer rate (\%) $\downarrow$ & 50.1\phantom{0} & 32.4\phantom{0} & \markeddelta{deltaBlueBackground}{-17.7\phantom{0}} & 26.8\phantom{0} & 27.7\phantom{0} & \markeddelta{red!12}{+0.9\phantom{0}} \\
\bottomrule
\end{tabular}
\end{table}

We further analyze reports revised by the rule bank (Table~\ref{tab:changed-reports-full}). Draft and revised reports are evaluated on the same cases using the same reference reports and evaluation settings. For SelfCue-KD, the largest improvement among CT-FineBench attribute groups is on size questions (+33.2 points), alongside a 17.7-point reduction in the no-answer rate. For NV-Reason-CT, scores for size, morphology, and density improve, while those for location and distribution decline. Rule bank revision improves size-related reporting for both report models. However, the lower location and distribution scores for NV-Reason-CT suggest that some revisions may improve certain finding attributes while compromising others.

\subsection{Learned Rules}
\label{app:learned-rules}

The final rule banks contain 30 entries for SelfCue-KD and 13 for NV-Reason-CT. Figure~\ref{fig:learned-rules} reproduces three entries covering a classifier-based addition, a measurement-based removal, and a measurement addition. Each entry stores the target finding, edit type, evidence condition, editing instruction, and an explanation of the evidence relationship. The structured evidence condition specifies when a rule applies, while the editing instruction describes the local revision and how to handle related statements in the draft. For example, the cardiomegaly addition addresses existing negative statements and avoids repeated additions. The pericardial-effusion removal instructs the editor agent to preserve other findings in the same sentence. These instructions constrain how an applicable rule is executed, beyond identifying the target abnormality.

\begin{figure*}[t]
\centering
\lstset{style=rulebankentry}

{\small\textbf{SelfCue-KD (R13)}\par}
\begin{lstlisting}
{
  "id": "R13",
  "finding": "Cardiomegaly",
  "edit": "add",
  "when_code": [["clf:Cardiomegaly", ">=", 0.7749]],
  "do": "Add the sentence 'The heart size is increased, consistent with cardiomegaly.' to the report: this rule applies even when no sentence mentions Cardiomegaly. If sentences deny Cardiomegaly, none may do so any more: a sentence that denies only that is removed, except that the first such sentence is replaced with the sentence above; a sentence that denies other things too keeps those and drops only Cardiomegaly. The sentence above appears exactly once in the report.",
  "why": "The classifier probability directly estimates the presence of cardiomegaly, so a high value provides direct evidence supporting the addition of this finding."
}
\end{lstlisting}

{\small\textbf{SelfCue-KD (R4)}\par}
\begin{lstlisting}
{
  "id": "R4",
  "finding": "Pericardial effusion",
  "edit": "remove",
  "when_code": [["pericardial_ml", "<=", 15.8806]],
  "do": "The report must no longer state Pericardial effusion. Remove that statement everywhere, also inside list sentences and in the impression. When a sentence states other findings too, keep them and drop only this one; delete a sentence only when nothing else in it remains.",
  "why": "The measurement pericardial_ml directly quantifies fluid in the pericardial sac, which is the defining characteristic of a pericardial effusion. A low volume (≤ the threshold) indicates little or no fluid, supporting the removal of a stated pericardial effusion finding. The evidence is directly related to the finding and the direction (low volume → remove) is appropriate."
}
\end{lstlisting}

{\small\textbf{NV-Reason-CT (R3)}\par}
\begin{lstlisting}
{
  "id": "R3",
  "finding": "ascending aorta",
  "edit": "add_value",
  "when_code": [["ascending_mm", ">=", 40.45]],
  "do": "Add '<value sentence>' to the findings.",
  "why": "The evidence is a direct measurement of the ascending aorta (ascending_mm), which is the same structure as the finding. A large diameter (>= threshold) is relevant to reporting the aortic size, so the direction (adding the measured value when the measurement is high) is appropriate. No unrelated structure or reversed direction is involved."
}
\end{lstlisting}

\caption{Selected rule-bank entries in their original wording. Development performance records are omitted.}
\label{fig:learned-rules}
\end{figure*}

\section{Independent Evidence Analysis}
\label{app:independent-evidence-analysis}

\subsection{Rules Learned from Drafts Alone}
\label{app:draft-only-rules}

We inspect draft-only rules to understand what an agent learns without independent image evidence (Figure~\ref{fig:draft-only-rules}). Before post-learning checks, the banks contain eight rules for SelfCue-KD and five for NV-Reason-CT. The learning agent receives revision-outcome feedback but constructs rule conditions from report length, abnormality counts, and RadBERT scores extracted from the draft.

\begin{figure*}[t]
\centering
\lstset{style=rulebankentry}

\textbf{SelfCue-KD (draft-only, R5)}\par
\begin{lstlisting}
{
  "id": "R5",
  "finding": "Atelectasis",
  "edit": "add",
  "when_code": [["draft_words", ">=", 328.24]],
  "do": "Add the sentence 'There are linear atelectasis in both lungs.' to the report: this rule applies even when no sentence mentions Atelectasis. If sentences deny Atelectasis, none may do so any more: a sentence that denies only that is removed, except that the first such sentence is replaced with the sentence above; a sentence that denies other things too keeps those and drops only Atelectasis. The sentence above appears exactly once in the report.",
  "why": "new evidence-direction combo"
}
\end{lstlisting}

\textbf{SelfCue-KD (draft-only, R7)}\par
\begin{lstlisting}
{
  "id": "R7",
  "finding": "Lung nodule",
  "edit": "add",
  "when_code": [["draft:Hiatal hernia", ">=", 0.0061]],
  "do": "Add the sentence 'There are millimetric nonspecific nodules in both lungs.' to the report: this rule applies even when no sentence mentions Lung nodule. If sentences deny Lung nodule, none may do so any more: a sentence that denies only that is removed, except that the first such sentence is replaced with the sentence above; a sentence that denies other things too keeps those and drops only Lung nodule. The sentence above appears exactly once in the report.",
  "why": "new evidence for lung nodule add"
}
\end{lstlisting}

\caption{Selected draft-only rules before post-learning checks. Development performance records are omitted.}
\label{fig:draft-only-rules}
\end{figure*}

The examples link report length to atelectasis and a draft-derived hiatal-hernia score to lung nodules. Such associations may fit development cases without providing image evidence for the target abnormality. Some draft-derived signals remain informative, but the resulting revision gains are limited. Even without medical plausibility review, CE-F1 improves by only 0.21 points for SelfCue-KD and 0.33 points for NV-Reason-CT, compared with 11.50 and 0.89 points when independent image evidence is available. These results support the value of new image observations for learning effective revision rules.

\subsection{Classifier and Report Performance}
\label{app:classifier-report-performance}

We compare the independent image classifiers with drafts and revised reports to examine their complementary strengths in abnormality detection (Table~\ref{tab:classifier-report-performance}). The classifiers have higher recall than either report model's drafts, while the drafts have higher precision. FlexiCT has higher standalone CE-F1 than the three-classifier mean. Rule bank revision substantially improves abnormality detection in SelfCue-KD. It also improves CE-F1 in NV-Reason-CT, even though the independent classifiers have lower standalone CE-F1 than this report model. These results demonstrate the value of complementary image evidence for improving an already strong report model.

\begin{table}[t]
\caption{Classifier and report performance on 1,564 test scans. Precision, recall, CE-F1, and AUROC are macro-averaged over the 18 abnormalities.}
\label{tab:classifier-report-performance}
\centering\scriptsize
\setlength{\tabcolsep}{1.8pt}
\renewcommand{\arraystretch}{1.12}
\begin{tabular}{@{}p{0.36\linewidth}rrrrr@{}}
\toprule
Model & Precision & Recall & CE-F1 & CRG & AUROC \\
\midrule
\multicolumn{6}{l}{\textbf{Image classifiers}} \\
FlexiCT & 57.26 & 67.02 & 61.22 & 54.77 & 87.02 \\
HLIP & 48.19 & 64.71 & 54.16 & 51.11 & 82.89 \\
ASAP & 46.80 & 69.98 & 55.38 & 52.26 & 83.29 \\
Three-classifier mean & 53.54 & 68.86 & 59.73 & 54.47 & 86.52 \\
\midrule
\multicolumn{6}{l}{\textbf{SelfCue-KD}} \\
Draft & 57.72 & 43.47 & 46.20 & 43.86 & -- \\
+ RuleBank & 58.78 & 58.95 & 57.70 & 51.87 & -- \\
\midrule
\multicolumn{6}{l}{\textbf{NV-Reason-CT}} \\
Draft & 67.30 & 59.93 & 63.03 & 53.97 & -- \\
+ RuleBank & 66.95 & 61.92 & 63.91 & 54.94 & -- \\
\bottomrule
\end{tabular}
\end{table}

\section{Additional Ablation Studies and Analysis}
\label{app:additional-ablations}

\subsection{Fixed Clinical Rubric}
\label{app:fixed-measurement-thresholds}

We construct a simplified fixed clinical rubric using literature-informed measurement cutoffs (Table~\ref{tab:fixed-measurement-thresholds}). The cutoffs are fixed before evaluation and shared by both report models. A finding is added when its threshold condition is satisfied. On the opposite side of the cutoff, an existing statement is removed or rewritten as normal for aortic dilation. Using these criteria to remove or normalize an existing statement is an additional assumption of this rubric.

\begin{table}[t]
\caption{Measurement cutoffs and source excerpts for the fixed clinical rubric.}
\label{tab:fixed-measurement-thresholds}
\centering\scriptsize
\setlength{\tabcolsep}{2.5pt}
\renewcommand{\arraystretch}{1.15}
\begin{tabular}{@{}p{0.24\linewidth}p{0.27\linewidth}p{0.43\linewidth}@{}}
\toprule
Target abnormality & Measurement and cutoff & Source and supporting text \\
\midrule
\raggedright Aortic dilation & \raggedright Ascending aorta diameter\newline $\geq 40$ mm & \raggedright \citep{munden2018managing}: ``2 SD above the mean diameter as adjusted for age, sex, and body surface area''.\newline Our shared 40 mm cutoff is adapted from population measurements \citep{rogers2013distribution}. \tabularnewline
\addlinespace[3pt]
\raggedright Coronary artery wall calcification & \raggedright Coronary calcium volume\newline $>0$ mm$^3$ & \raggedright \citep{hecht20172016}: ``CAC should be evaluated and reported on all noncontrast chest CT examinations''.\newline Presence is mapped to nonzero volume. \tabularnewline
\addlinespace[3pt]
\raggedright Hepatic steatosis & \raggedright Median liver attenuation\newline $\leq 40$ HU & \raggedright \citep{boyce2010hepatic}: ``moderate-to-severe hepatic steatosis (defined by liver attenuation $\leq 40$ HU)''. \tabularnewline
\addlinespace[3pt]
\raggedright Cardiomegaly & \raggedright Cardiothoracic ratio\newline $\geq 0.50$ & \raggedright \citep{061b830d88764658917332b3c0a2748e}: Radiograph criterion: ``If the CTR is greater than 0.50, cardiomegaly is assumed in the adult patient''. The CT cutoff ``was found to be 0.53''.\newline We apply the radiograph value to CT. \tabularnewline
\addlinespace[3pt]
\raggedright Arterial wall calcification & \raggedright Aortic wall calcium volume\newline $>0$ mm$^3$ & \raggedright \citep{hecht20172016}: ``It may be reasonable to evaluate and report thoracic aortic calcification''.\newline Nonzero aortic calcium is our proxy for the broader label. \tabularnewline
\bottomrule
\end{tabular}
\end{table}

\subsection{Learning Curves across Metrics}
\label{app:learning-metrics}

We provide exact scores at learning iterations 1 and 10 to supplement the learning curves (Table~\ref{tab:learning-metrics}). For SelfCue-KD, all three agent combinations improve CRG and CT-FineBench, accompanied by a small decrease in GREEN. NV-Reason-CT shows smaller changes across these metrics. Despite differences at the first iteration, the combinations reach similar CRG and CT-FineBench scores by iteration 10, indicating consistent revision performance across the tested agent implementations.

\begin{table}[t]
\caption{Revision scores at learning iterations 1 and 10, shown from left to right.}
\label{tab:learning-metrics}
\centering\scriptsize
\setlength{\tabcolsep}{2pt}
\renewcommand{\arraystretch}{1.12}
\begin{tabular}{@{}p{0.30\linewidth}rrr@{}}
\toprule
\shortstack[l]{Learning agent /\\editor agent} & CRG & GREEN & CT-FineBench \\
\midrule
\multicolumn{4}{l}{\textbf{SelfCue-KD}} \\
gpt-oss / gpt-oss & $47.52\to 51.91$ & $50.23\to 49.64$ & $36.07\to 41.38$ \\
Sonnet / gpt-oss & $45.40\to 51.99$ & $50.43\to 49.99$ & $29.10\to 41.28$ \\
Sonnet / Sonnet & $45.39\to 52.03$ & $50.44\to 50.05$ & $29.07\to 41.29$ \\
\midrule
\multicolumn{4}{l}{\textbf{NV-Reason-CT}} \\
gpt-oss / gpt-oss & $54.69\to 54.94$ & $45.60\to 45.93$ & $45.23\to 44.70$ \\
Sonnet / gpt-oss & $54.14\to 54.91$ & $45.73\to 45.70$ & $44.84\to 44.74$ \\
Sonnet / Sonnet & $54.14\to 54.89$ & $45.73\to 45.64$ & $44.84\to 44.73$ \\
\bottomrule
\end{tabular}
\end{table}

\subsection{Results across Draft Samples}
\label{app:draft-variation}

We provide per-seed draft and revised scores to supplement the draft-variation analysis (Table~\ref{tab:draft-variation-full}). Both report models improve CE-F1 across all five sampled draft sets. SelfCue-KD maintains substantial gains, while NV-Reason-CT gains range from 0.14 to 0.93 points. These results indicate that revision benefits persist across draft samples, although their magnitude depends on the sampled drafts.

\begin{table}[t]
\caption{Draft and revised scores for five sampling seeds. A separate rule bank is learned for each seed.}
\label{tab:draft-variation-full}
\centering\scriptsize
\setlength{\tabcolsep}{2pt}
\renewcommand{\arraystretch}{1.12}
\begin{tabular}{@{}lrrrrrrr@{}}
\toprule
 & & \multicolumn{3}{c}{CE-F1} & \multicolumn{3}{c}{GREEN} \\
\cmidrule(lr){3-5}\cmidrule(lr){6-8}
Seed & \shortstack{Changed\\(\%)} & Draft & Revised & $\Delta$ & Draft & Revised & $\Delta$ \\
\midrule
\multicolumn{8}{l}{\textbf{SelfCue-KD}} \\
1001 & 81.6 & 45.99 & 55.84 & \markeddelta{deltaBlueBackground}{+9.85} & 50.04 & 49.94 & \markeddelta{red!12}{-0.10} \\
2002 & 82.0 & 46.31 & 57.71 & \markeddelta{deltaBlueBackground}{+11.40} & 50.51 & 50.21 & \markeddelta{red!12}{-0.30} \\
3003 & 86.4 & 45.78 & 57.01 & \markeddelta{deltaBlueBackground}{+11.23} & 49.73 & 49.69 & \markeddelta{red!12}{-0.04} \\
4004 & 83.1 & 46.24 & 57.35 & \markeddelta{deltaBlueBackground}{+11.12} & 50.02 & 50.11 & \markeddelta{deltaBlueBackground}{+0.09} \\
5005 & 84.0 & 45.94 & 55.70 & \markeddelta{deltaBlueBackground}{+9.76} & 50.43 & 50.47 & \markeddelta{deltaBlueBackground}{+0.05} \\
\addlinespace[3pt]
\multicolumn{8}{l}{\textbf{NV-Reason-CT}} \\
1001 & 27.4 & 62.65 & 63.44 & \markeddelta{deltaBlueBackground}{+0.79} & 45.47 & 45.71 & \markeddelta{deltaBlueBackground}{+0.24} \\
2002 & 19.4 & 62.26 & 63.17 & \markeddelta{deltaBlueBackground}{+0.91} & 45.31 & 45.28 & \markeddelta{red!12}{-0.02} \\
3003 & 24.9 & 62.45 & 62.58 & \markeddelta{deltaBlueBackground}{+0.14} & 46.11 & 45.90 & \markeddelta{red!12}{-0.22} \\
4004 & 24.4 & 62.10 & 62.94 & \markeddelta{deltaBlueBackground}{+0.84} & 45.37 & 45.84 & \markeddelta{deltaBlueBackground}{+0.47} \\
5005 & 23.9 & 62.26 & 63.18 & \markeddelta{deltaBlueBackground}{+0.93} & 45.50 & 45.52 & \markeddelta{deltaBlueBackground}{+0.03} \\
\bottomrule
\end{tabular}
\end{table}

\subsection{Quantitative Information in Revised Reports}
\label{app:quantitative-information}

We examine added measurements by vessel and their agreement with evidence and reference reports to supplement the main quantitative-reporting analysis (Table~\ref{tab:quantitative-completion}). Values are extracted from report text using their units and anatomical sites, and reference comparisons use the largest diameter reported for the same vessel segment.

\begin{table}[t]
\caption{Added diameters and agreement with reference reports. MAE uses additions with matching reference measurements.}
\label{tab:quantitative-completion}
\centering\scriptsize
\setlength{\tabcolsep}{3pt}
\renewcommand{\arraystretch}{1.12}
\begin{tabular}{@{}p{0.44\linewidth}rrr@{}}
\toprule
Vessel & Added & \shortstack{Reference\\pairs} & MAE (mm) \\
\midrule
\multicolumn{4}{l}{\textbf{SelfCue-KD}} \\
Ascending aorta & 240 & 62 & 2.3 \\
Aortic arch & 0 & 0 & -- \\
Descending aorta & 260 & 12 & 2.7 \\
Main pulmonary artery & 17 & 5 & 1.0 \\
\midrule
\multicolumn{4}{l}{\textbf{NV-Reason-CT}} \\
Ascending aorta & 41 & 23 & 2.4 \\
Aortic arch & 0 & 0 & -- \\
Descending aorta & 0 & 0 & -- \\
Main pulmonary artery & 0 & 0 & -- \\
\bottomrule
\end{tabular}
\end{table}

All added diameters match the supplied measurements after rounding, and 557 of 558 additions name the vessel specified by the triggering rule. Reference comparisons use a selected subset with reported diameters; these pairs can overlap across report models. Their MAE describes agreement with reference text, while added values can inherit systematic measurement differences. Revision also affects existing numerical content: SelfCue-KD removes 127 values, including 67 through aortic normalization, whereas NV-Reason-CT removes none. These findings support traceable numerical completion.

\section{Edit Assessment Details}
\label{app:case-analysis}

Claude Sonnet 5.5 assesses all edits from the gpt-oss-120b editor agent on the 1,564 test scans using reference reports and exact text changes, without image input. Labels are supported, partly supported, not mentioned, or contradicted. Partial support indicates a matching abnormality with a different attribute or value. Removals require explicit reference statements of absence or normality to be supported. Diameter support requires agreement within 3 mm of a reference value or agreement with a qualitative normal/dilated description, using matching limits of 40 mm for the ascending aorta and 30 mm for the descending aorta and pulmonary artery. Three Claude Sonnet 5.5 prompt variants rate the potential importance of 43 revision types, assuming correctness. Their median ratings are none, minor, moderate, or major. Classifier groups use classifier conditions; measurement groups use measurement conditions or supply measured values or attributes. Percentages are within-group shares, and the All edits rows include edits triggered by the draft alone.

\end{document}